\documentclass[lettersize,journal]{IEEEtran}
\usepackage{amsmath,amsfonts}
\usepackage{algorithmic}
\usepackage{algorithm}
\usepackage{array}
\usepackage[caption=false,font=normalsize,labelfont=sf,textfont=sf]{subfig}
\usepackage{textcomp}
\usepackage{stfloats}
\usepackage{url}
\usepackage{verbatim}
\usepackage{graphicx}
\usepackage{cite}
\usepackage{enumitem}
\begin{document}

\title{CausalChat: Interactive Causal Model Development and Refinement Using Large Language Models}

\author{Yanming Zhang, Akshith Kota, Eric Papenhausen, and Klaus Mueller,~\IEEEmembership{Fellow, IEEE}
\thanks{All authors are with the Computer Science Department at Stony Brook University. Contact author: yanming.zhang@stonybrook.edu }
\thanks{Manuscript received September 19, 2024; revised xxx}}

\markboth{Journal of \LaTeX\ Class Files,~Vol.~14, No.~8, August~2024}%
{Shell \MakeLowercase{\textit{et al.}}: A Sample Article Using IEEEtran.cls for IEEE Journals}


\maketitle

\begin{abstract}
   Causal networks are widely used in many fields to model the complex relationships between variables. A recent approach has sought to construct causal networks by leveraging the wisdom of crowds through the collective participation of humans. While this can yield detailed causal networks that model the underlying phenomena quite well, it requires a large number of individuals with domain understanding. We adopt a different approach: leveraging the causal knowledge that large language models, such as OpenAI's GPT-4, have learned by ingesting massive amounts of literature. Within a dedicated visual analytics interface, called CausalChat, users explore single variables or variable pairs recursively to identify causal relations, latent variables, confounders, and mediators, constructing detailed causal networks through conversation.
   Each probing interaction is translated into a tailored GPT-4 prompt and the response is conveyed through visual representations which are linked to the generated text for explanations. We demonstrate the functionality  of CausalChat across diverse data contexts and conduct user studies involving both domain experts and laypersons.
\end{abstract}


\section{Introduction}
\IEEEPARstart{C}{ausal} relationships are the building blocks of how we make sense of the world. They help us understand why things happen the way they do, from the simple cause-and-effect of a light switch to the complex interplay of factors influencing societal trends. We encounter causal facts in everyday life, whether in conversations about health choices or discussions on broader issues like health policies.

Beneath this universal concept lies a deep philosophical debate between 18th-century thinkers David Hume and Immanuel Kant. Hume, an empiricist, viewed causality as a mental habit formed by repeated experiences \cite{hume2007enquiry}, while Kant, a transcendental idealist, saw it as an inherent concept imposed by the mind on sensory experience \cite{kant1908critique}.
The Hume-Kant debate echoes in the modern discussion of causal inference from data versus text. Data-driven methods, like Hume's empiricism, rely on statistical analysis of observations, while text-driven methods, aligned with Kant's idealism, extract causal knowledge from literature or people. Both have limitations: data-driven methods struggle with unobserved confounders, biases, and overfitting \cite{pearl2016causal}, while text-based methods face challenges like language ambiguity, lack of domain expertise, and difficulty distinguishing causation from correlation \cite{pearl2019seven}.


In this paper, we integrate perspectives from Kant and Hume in advancing causal analysis. To incorporate Kant's perspective we leverage large language models (LLMs) trained on extensive text data, and so harness their rich causal knowledge to minimize ambiguities. Following Hume's approach, we also incorporate data when available. 
While this combined strategy is not entirely new \cite{yen2023crowdidea}, to our knowledge this approach has not been explored thus far with LLMs. It allows users to navigate and aggregate complex causal relationships, enhancing accessibility even for those without domain expertise.

A widely known issue with LLMs like ChatGPT is that they can produce inaccurate information and hallucinations, even with well-constructed prompts \cite{kiciman2023causal}. To address this problem, we approach causal questions from every possible angle, enhancing confidence in selected relationships while uncovering potential conflicts. A useful aspect of posing causal questions from diverse perspectives is that it sheds light on the broader context of the relationship. ChatGPT often highlights latent, confounding and mediator variables, which offer valuable insights. Taking an additional step to explicitly request these variables facilitates domain knowledge acquisition, supporting focused data collection and refinement of causal models.

Finally, while multifaceted prompts like these offer valuable perspectives, they generate substantial textual data, which can potentially be overwhelming even with summarization. To aid human analysts in navigating this information, we introduce a set of interactive visualizations designed to efficiently convey  key insights. This paper presents our approach and its application in developing and refining causal networks from text and data, involving active human participation.

In summary, our contributions are as follows:
\begin{itemize}
  \item a practical implementation of a recursive text/data-driven causal network development and refinement paradigm
  \item a set of LLM prompts that interrogate a hypothesized causal relation from diverse perspectives, such as “does high A cause low B” and so forth
  \item a suite of visual representations and charts to communicate the relation’s diverse causal perspectives and its latent variables, confounders, and mediators
  \item a visual component capable of managing feedback loops in causal inference, extending into multiple causal networks tailored for diverse decision-making processes
  \item an explainable AI mechanism that links the visual representations back to the prompted LLM text
  \item an interactive visual interface that puts the analyst into the loop of the recursive text/data-driven causal network development and refinement activities 
  \item a set of case studies and usage scenarios that demonstrate our ideas and studies with experts and laypersons to gain insight about the usage of our system    
\end{itemize}

Fig. \ref{fig:teaser} shows our visual analytics dashboard with all functional components for a causal model of a car. 
In the following we will first describe relevant related work (Section 2), our methodology (Section 3), some usage scenarios (Section 4), our user study (Section 5), and a discussion (Section 6). Preliminary results of this work have appeared in\cite{zhang2023explainable}.

\section{Related Work}

Causal networks are widely used in various fields, such as epidemiology \cite{vandenbroucke2016causality}, healthcare \cite{glass2013causal}, biology \cite{dang2015reactionflow}, and social sciences \cite{gerring2005causation} to understand complex systems. In observational studies, causal networks are typically represented using directed acyclic graphs (DAGs), as DAGs can illustrate potential sources of confounding bias and selection bias\cite{murray2022wheel,greenland1999causal}. Constructing such a DAG is best achieved through the rigorous process of randomized controlled trials based on first principles. However, this method often faces challenges, be it due to cost, ethical considerations, or practical limitations. Additionally, it is  not easily scalable and may restrict the number of researchers who can engage in such studies. Alternatively, a more scalable and general approach involves deriving a causal DAG through one of three primary methods: analyzing data, analyzing text, or collaborative construction and crowdsourcing. Many of these approaches combine elements from more than one of these paradigms, and some allow human analysts to participate in the DAG development process \cite{wang2015visual, wang2017visual, xie2020visual}.

\subsection{Causal Network Discovery using Numerical Data}
There are essentially two popular strategies for causal discovery. One approach involves enforcing the constraint that two statistically independent variables are not causally linked, followed by a series of conditional independence tests to construct a compliant DAG. Well-known algorithms for this method include the PC algorithm \cite{spirtes2000causation} and the Fast Causal Inference (FCI) algorithm \cite{spirtes2001anytime}. Another strategy is to greedily explore the space of possible DAGs via Greedy Equivalence Search (GES) \cite{chickering2002optimal}. This entails score-based methods in which edges are iteratively added and removed from the graph to maximize a model fitness measure, such as the Bayesian Information Criterion (BIC) \cite{burnham2004multimodel, schwarz1978estimating}.

Causal discovery relies on four common assumptions: (1) the causal structure can be represented by a DAG, (2) all nodes are Markov-independent from non-descendants when conditioned on their parents, (3) the DAG is faithful to the underlying conditional independence, and (4) the DAG is sufficient, i.e., there is no pair of nodes with a common external cause. Unfortunately, these conditions are rarely entirely met, often due to selection/sampling bias in the data. 
Essentially, the phenomenon to be explained by the causal network is only partially captured by (1) the measured variables and (2) the observed data samples, and this leads the discovery algorithm astray. While the probability of obtaining a partially incorrect DAG can be reduced by using more data, it remains uncertain how much data is truly needed \cite{Shalizi-chapter25, robins2003uniform}. 

Our research tackles both of these bottlenecks: (1) the limitation of collected datasets in fully capturing all variables essential for constructing a comprehensive causal model, which we address through ChatGPT-based variable ideation and relevance assessment, and (2) in the absence of data for newly discovered relations connecting ideated variables to the DAG, we use ChatGPT to generate plausible hypotheses for the strengths of these links based on its contextual understanding and patterns learned from vast sources.


\subsection{Causal Network Discovery using Textual Data}

While extracting causal relations from text documents is not a new endeavor \cite{yang2022survey}, thanks to ChatGPT this process has become remarkably convenient with a simple prompt across various application domains. The literature on LLM assisted causal network learning is rapidly expanding. The earliest documented attempt using LLMs (specifically GPT-3) for causal analytics was by Long et al. \cite{long2023langcaus}. However, this was a preliminary study focused on optimizing prompts to reveal insights into the presence or absence of directed edges. More recently, Kıcıman et al. \cite{kıcıman2023causal} delved much deeper into the subject. They devised a comprehensive set of prompts for GPT-3.5 and GPT-4, generating yes/no responses to standard causal queries. While they demonstrated excellent success rates on benchmark datasets where causal truth was known, they did not explore utilizing GPT's output to gather additional causal and contextual knowledge. Similarly, 
subsequent papers (e.g., \cite{nam2023show, long2023causal, gao2023chatgpt}) also did not explore visualizing the acquired information within explainable and trustworthy AI.

Some studies have highlighted the limitations of using LLMs for causality analysis. While LLMs excel at discerning causality from empirical or commonsense knowledge, Jin et al. \cite{jin2023can} demonstrated their significantly reduced effectiveness in deriving causality through pure causal reasoning—something numerical algorithms like PC are specifically designed for. To test this, they assembled a substantial dataset comprising over 400,000 correlational statements in natural language and tasked the LLM with determining the causal relationship between variables. Their findings revealed that existing LLMs demonstrated performance is akin to random chance in this particular task. These findings are echoed by Zečević et al. \cite{zevcevic2023causal}, who suggest that LLMs can serve as a valuable starting point for learning and inference, reaffirming their role as a tool for ideation and creativity. They can complement data-driven causal inference methods, such as PC, which is what one of the approaches we  promote here in this paper.

\subsection{Collaborative Causal Network Discovery with Crowds}

In 2018, Berenberg and Bagrow \cite{berenberg2018efficient} introduced a methodology that harnessed the 'wisdom of the crowds' to construct a large causal network, utilizing the widely-used crowdsourcing platform Amazon Mechanical Turk. They devised a three-stage approach: in stage 1, workers proposed causes; in stage 2, they suggested effects for these causes; and in stage 3, they edited and refined longer causal pathways derived from the stage 2 results. The final causal network was then formed by amalgamating all worker-generated pathways, with more popular edges indicating stronger causal links. Salim et al. \cite{salim2024belief} adopted a similar approach,  focusing on mining crowd beliefs and misconceptions in complex systems with societal impact such as climate change.

It is worth noting that the study by Berenberg and Bagrow predates the emergence of LLMs. While the degree to which Large Language Models (LLMs), trained on extensive human-written text, tap into the 'wisdom of the crowds' remains uncertain, it is plausible to expect that LLM assistance would necessitate a significantly smaller crowd. As LLMs effectively encapsulate the viewpoints of a large crowd simultaneously, using a few-shot prompt programming approach can guide and constrain the response towards a pertinent answer \cite{reynolds2021prompt}. We believe that our multifaceted prompt represents a significant step in this direction.

Yen et al. \cite{yen2021narratives+} developed an interactive system for collaborative causal network construction. This system enabled users to articulate narratives to explain causal relationships they perceived, visualize the causal models of these using DAGs, and review and incorporate the causal diagrams and narratives of other users. A notable feature of their system was the 'Inspire Me' popup, which users could request when they needed fresh ideas on how to expand the network. They would then be presented with one of several pre-programmed thought-provoking questions related to causal relationships.

The purpose of this system was to investigate whether actively evolving and narrating a causal network, and learning from networks constructed by peers, could uncover blind spots in a person’s causal reasoning and lead to a refinement of their own causal network. In a more recent development, the authors introduced an enhanced user interface called CrowdIDEA \cite{yen2023crowdidea}. This version also included a data panel with visualizations and statistics. It is conceivable that an LLM could fulfill a similar collaborative role. For instance, in our system, users have the ability to explore any variable or variable pair and visualize suggested directions, confounders, and mediators, all ready to be seamlessly integrated into the emerging causal network.  

\begin{figure}[t!]
\centering
\includegraphics[width=\linewidth]{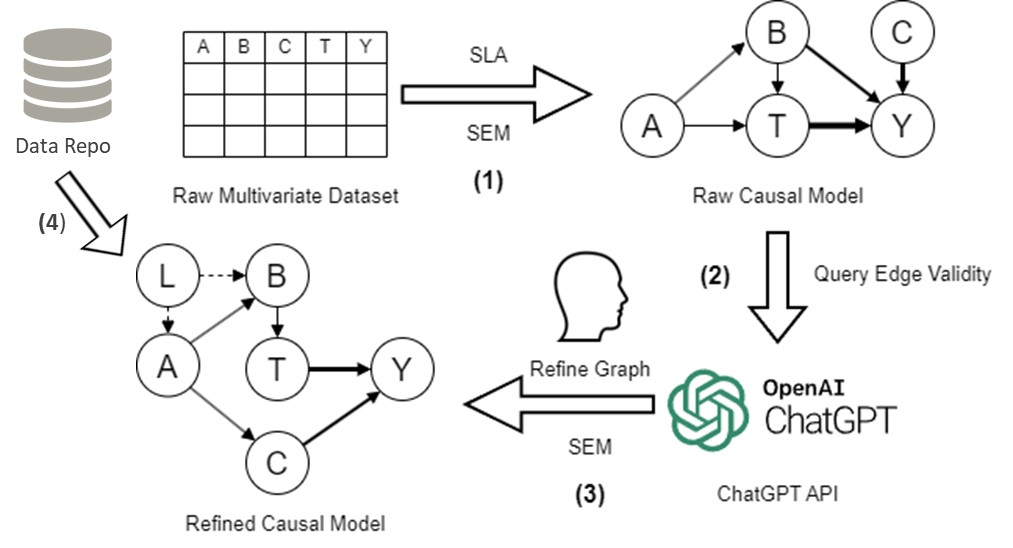}
\vspace{-15pt}
\caption{Workflow of our ChatGPT-powered Causal Auditor. (1) (Optional) algorithmic discovery of the initial (raw) causal model. (2) Query-driven ChatGPT-based edge commentary. (3) Analyst-initiated model refinement informed by the outcomes of steps 1 and 2. (4) Data upload for newly introduced variables and relations (if available). (SLA stands for Structure Learning Algorithm and SEM stands for Structural Equation Modeling).}
\label{fig:workflow}
\vspace{-15 pt}
\end{figure}

\section{Methodology}

Fig. \ref{fig:workflow} illustrates the workflow of our method. Step 1 computes an initial causal model from the user-provided data. This model is displayed as a Completed Partially Directed Acyclic Graph (CPDAG) in a our visual causal analytics interface, where the model coefficients are computed through Structural Equation Modelling (SEM). Alternatively, users may skip this data-driven step and draw their own causal network. The model can then be refined based on the outcome of a ChatGPT prompt with regards to a certain relation indicated by the user -- we have used the GPT-4 API which at the time of this research was the most advanced OpenAI GPT version. In the remainder of this paper will refer to this API as 'GPT-4'. 

To aid in the GPT-4 powered causal analytics activity, we provide various visualizations that summarize the gist of these responses, linked back to the corresponding GPT-4 response text. Finally, if a new variable or relation is introduced, the respective data may be uploaded (if available) from a suitable data repository and the model coefficients are updated. The workflow adopts a loop optimization structure, wherein the incorporation of new data and the updating of model coefficients aligns with Hume's empiricism, while the inferred causal knowledge provided by GPT-4 is consistent with Kant's epistemology.

We investigate causality at two levels: the first examines the existence of direct causal relations, while the second digs deeper to uncover latent variables, mediators, and confounders.


\subsection{Prompting for Direct Causal Relationships}

The prompts we utilize follow an optimized template (see supplemental material). Offering sufficient contextual information and guidance on expectations is crucial for prompt engineering \cite{reynolds2021prompt, mishra2021reframing}. Below is an abstraction of the prompt.

\vspace{5pt}

\noindent \textbf{Prompt:} You are an expert in $<$domain$>$. On a scale from 1 to 4, where 4 represents highly significant, 3 represents significant, 2 represents doubtful, 1 represents not significant, rate the following cause-and-effect relationship: Does higher/lower {A/B} cause higher/lower {B/A}.   

\vspace{5pt}

\noindent This generates 10 distinct prompts, 5 each for A and B taking opposite roles, and within each of these two sets there are 4 combinations of A and B being (higher, lower) plus one relation that just asks this for a general case. An example prompt is shown below, where (...) denotes further prompt specifications (see supplement for complete prompt): 

\vspace{5pt}

\noindent \textbf{Prompt:} You are an expert in public health. On a scale from 1 to 4, where 4 represents highly significant, 3 represents significant, 2 represents doubtful, 1 represents not significant, rate the cause-and-effect relationship: Does \textit{higher percent fair or poor health} cause \textit{lower life expectancy}...

\vspace{5pt}

\noindent Including the domain hint 'public health'  provides contextual information. GPT-4 can also infer the domain from the dataset attributes if it is told to do so. 
GPT-4's response to the prompt is to the point:

\vspace{5pt}

\noindent \textbf{Response}: Rating: 4

\subsection{Prompting for Confounders}

This prompt template also distinguishes among the 4 combinations that explore the effects of higher and lower levels plus one relation that just asks this for the general case. In the following we explain this template using the variables \textit{food environment index} and \textit{violent crime rate} as an example. Also here (....) denotes omissions (see supplement).

\vspace{5pt}

\noindent \textbf{Prompt:} You are an expert in public health. Given the cause-and-effect relationship ‘lower food environment index’ causes ‘higher violent crime rate’ identify potential confounders based on the definition .... For each identified confounder, provide the following details in a tuple format:
1. Name of the confounder.
2. Strength of the confounder (options: weak, medium, strong).
3. Justification for its role as a confounder based on the definition provided.

\vspace{5pt}


\vspace{5pt}



\noindent \textbf{Response}:  GPT-4 returned 2 'strong' confounders (Socioeconomic Status, Residential Segregation) and 4 'medium' confounders (Substance Abuse and Mental Health Issues, Availability of Public Services, Racial and Ethnic Composition, Neighborhood Disorganization). For each a detailed justification was given, such as "Substance abuse and mental health issues can contribute to both a lower food environment index (due to prioritization of immediate needs over healthy food choices) and higher rates of violent crime, as these issues can lead to unstable social environments".

\subsection{Prompting for Mediators}

Also here we distinguished among the 4 level combinations and the general one. Using the same example as for the confounder, the (partial) mediator prompt is below (see supplement for complete prompt).  

\vspace{5pt}

\noindent \textbf{Prompt:} You are an expert in public health. Given the cause-and-effect relationship ‘lower food environment index’ causes ‘higher violent crime rate’ identify potential mediators based on the definition: Rather than a direct causal relationship between the independent variable and the dependent variable, the independent variable influences the mediator variable, which in turn influences the dependent variable. For each identified mediator, provide the following details in a tuple format:
1. Name of the mediator.
2. Strength of the mediator (options: weak, medium, strong).
3. Justification for its role as a mediator (...)
4. Specific conditions under which the mediator operates (...)
5. Direction of the mediator's effect ('positive' or 'negative') (...). The direction tells us how to intervene on the mediators to achieve the relationship....

\vspace{5pt} 

\noindent The 'Direction' parameter (parameter 5) is crucial as it specifies the way in which the level of a mediator should change to influence the effect variable as indicated. This guidance not only informs analysts about the type of intervention required but also sheds light on the underlying cause-effect mechanism.

\vspace{5pt} 

\noindent \textbf{Response}: GPT-4 returned 1 'strong' mediator (Economic Disadvantage $\uparrow$) and 4 'medium' mediators (Social Cohesion $\downarrow$, Substance Abuse $\uparrow$, Educational Attainment $\downarrow$, Mental Health $\downarrow$), where $\downarrow$ $\uparrow$ indicate the direction the mediator needs to have to support the effect. For each mediator a detailed justification was given, such as "A lower food environment index may contribute to reduced social cohesion within a community, as limited access to nutritious food options can lead to increased stress and poorer overall health. Reduced social cohesion has been associated with higher rates of violent crime, as it may lead to weaker community bonds and less effective informal social control".

\subsection{Prompting for Latent Factors}
\label{propmt_latent}
Unlike previous prompts, this prompt focuses on a single variable, emphasizing intervenable factors. It identifies actionable variables as points of intervention, enabling practitioners to influence the target variable through specific causal pathways. An example of a latent factors prompt is provided below (see supplement for the full prompt).

\vspace{5pt}

\noindent\textbf{Prompt:} Given the target variable \textit{primary care physicians rate}, identify potential latent (intervenable) factors that might influence the target variable. Ensure that the identified latent factors can be actionable or intervenable to affect the target variable. Provide the following details for each latent factor:
1. Name of the latent factor. 
2. Strength of the effect (weak, medium, strong).
3. Sign of the effect (positive, negative, or categorical).
4. Justification for its role as a latent factor.

\vspace{5pt}

\noindent\textbf{Response:} GPT-4 returned 1 'strong' positive latent factor (Reimbursement Rates), 2 'medium' positive latent factors (Medical Infrastructure Investment and Healthcare Policy Reforms),  1 'strong' negative latent factor (Medical Student Debt), and 1 'medium' negative latent factor (Urbanization Incentives). For each latent factor a storyline rationale was given. For instance, Medical Student Debt can serve as a negative latent factor, as 'high levels of debt from medical education can deter graduates from entering lower-paying specialties like primary care.' Medical Student Debt can be addressed through governmental medical debt relief programs, which act as intervention points.

\vspace{5pt}
 
\subsection{Visualizing the GPT-4 Generated Text Responses }

While the inclusion of text helps justify the presence (or absence) of a causal relation, GPT-4 may generate excessive text. Even when instructed to summarize its findings, this abundance of information can overwhelm general users. Below, we present the visualizations we have designed to make browsing this information easier.

 \vspace{5pt}

\subsection{The Causal Debate Chart}

\begin{figure}[t!]
\centering
\includegraphics[width=0.9\linewidth]{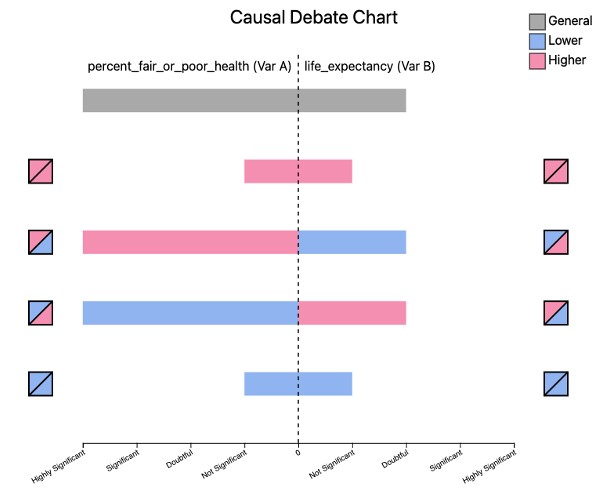}
\vspace{-5pt}
\caption{Causal Debate Chart for the relation \textit{Percent Fair or Poor Health - Life Expectancy}, presenting an overwhelming belief that the former is the cause of the latter.}
\label{fig:cs-dbt}
\end{figure}

Fig. \ref{fig:cs-dbt} shows our summary visualization that contrasts the numerical outcomes of the 10 prompts designed to probe a direct causal relation. We call it \textit{Causal Debate Chart} since it visually argues the strength of one variable being the cause of the other. The chart is a bidirectional bar chart where each side is headed by one of the two relation variables. In  this case the left side is \textit{Percent Fair or Poor Health (PFPH)} and the right side is \textit{Life Expectancy (LE)}. The x-axis is the score assigned by GPT-4 and the length of each bar is mapped to that score. The grey bars are for the general prompt while the other bars are colored in magenta if the cause was a higher or increasing level of the variable or in sky blue if the cause was a lower or decreasing level (see color legend on the top right). The textual level descriptions have been deliberately chosen to be relatable to humans. 

Let us now evaluate the chart. We observe that for the first (grey) set of bars \textit{PFPH} has a substantially longer bar (level 4) than \textit{LE} which has level 2 (level 2 is a doubtful cause in GPT-4 semantics). It means that the former wins the causal debate -- it has causal dominance. \textit{PFPH} seems to be a general cause of \textit{LE}.

Let's examine the other bars representing specific level studies. Here, we assess whether GPT-4 maintains consistent logic (as opposed to hallucinating). We observe that high or increasing \textit{PFPH} leads to low or decreasing \textit{LE} in the third set, and the same holds for the opposite relation in the fourth set. Sets two and five display low bars on both sides, as expected if the relation indicated by the other bars is considered true. The Causal Debate Chart in Fig. \ref{fig:cs-dbt} serves as a prime example of what we would expect from a steadfast causal relation.

\begin{figure}[t!]
\centering
\includegraphics[width=0.9\linewidth]{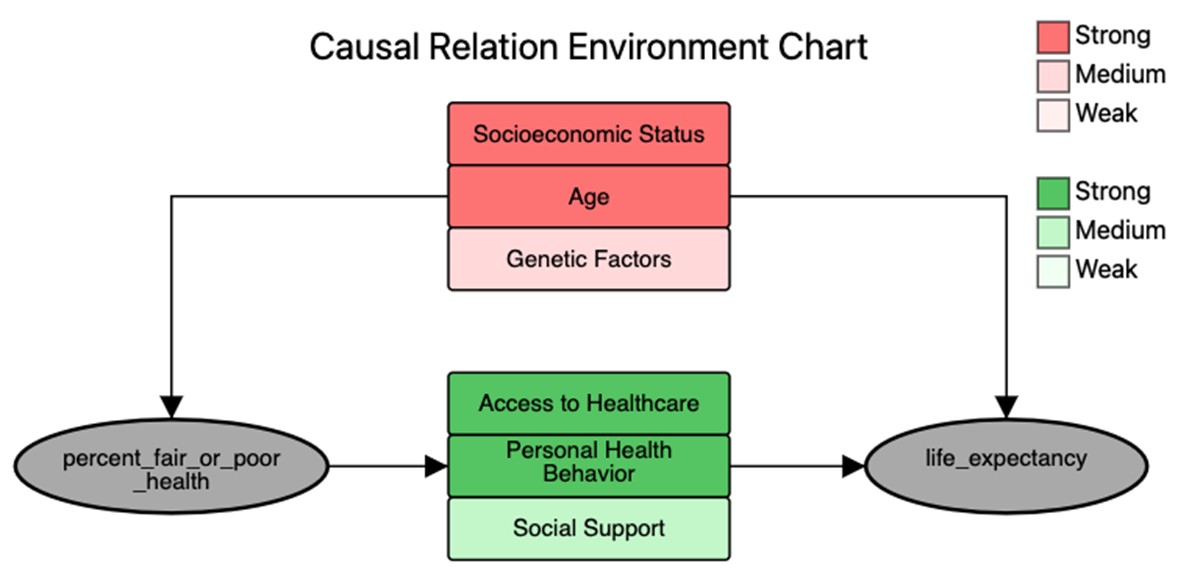}
\caption{Causal Relation Environment Chart for the relation \textit{Percent Fair or Poor Health - Life Expectancy}. The intensity of red and green encodes the strength of the mediators and covariates (weak, medium, strong), and the color of the cause and effect variables have the same interpretation as those in Fig. \ref{fig:cs-dbt}; in this specific case they are grey.}
\label{fig:cs-rel-env}
\vspace{-15 pt}
\end{figure}

\subsection{The Causal Relation Environment Chart}
The \textit{Causal Relation Environment Chart} is a diagram that visualizes the complete causal relation, consisting of the two main relational variables along with a list of mediators and confounders/covariates. An example for the general \textit{Percent Fair or Poor Health (PFPH) - Life Expectancy (LE)} relation is shown in Fig. \ref{fig:cs-rel-env}, where the confounders are colored in shades of red and the mediators are colored in shades of green -- the shades indicate their strength (see color legend on the right). 

\begin{figure}[t!]
\centering
\includegraphics[width=0.9\linewidth]{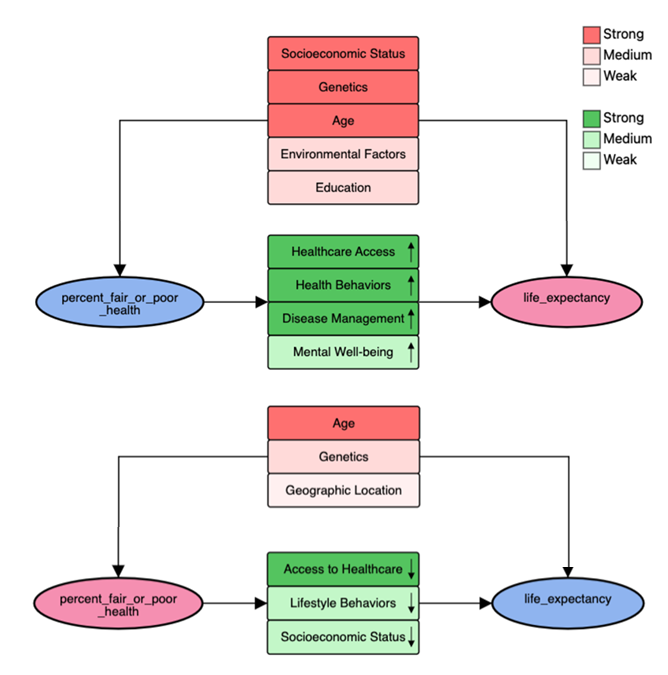}
\caption{Causal Relation Environment Chart for two level combinations of the relation \textit{Percent Fair or Poor Health - Life Expectancy}. The up and down arrows show the appropriate signs of the mediators.}
\label{fig:cs-rel-comb}
\end{figure}

It is often the case that GPT-4 will identify similar mediators and confounders for the more focused (low, high) relations. But they vary in the sign. For example,  to go from \textit{low PFPH} to \textit{high LE}, positive levels of the mediators are cited, such as good access to healthcare and good health habits, while to go from \textit{high PFPH} to \textit{low LE} the cited mediators are usually the opposite, like limited access to health care and poor health habits. Fig. \ref{fig:cs-rel-comb} shows these two cases where the up and down arrows indicate the positive and negative levels, respectively. 

\begin{figure}[t!]
\centering
\includegraphics[width=0.92\linewidth]{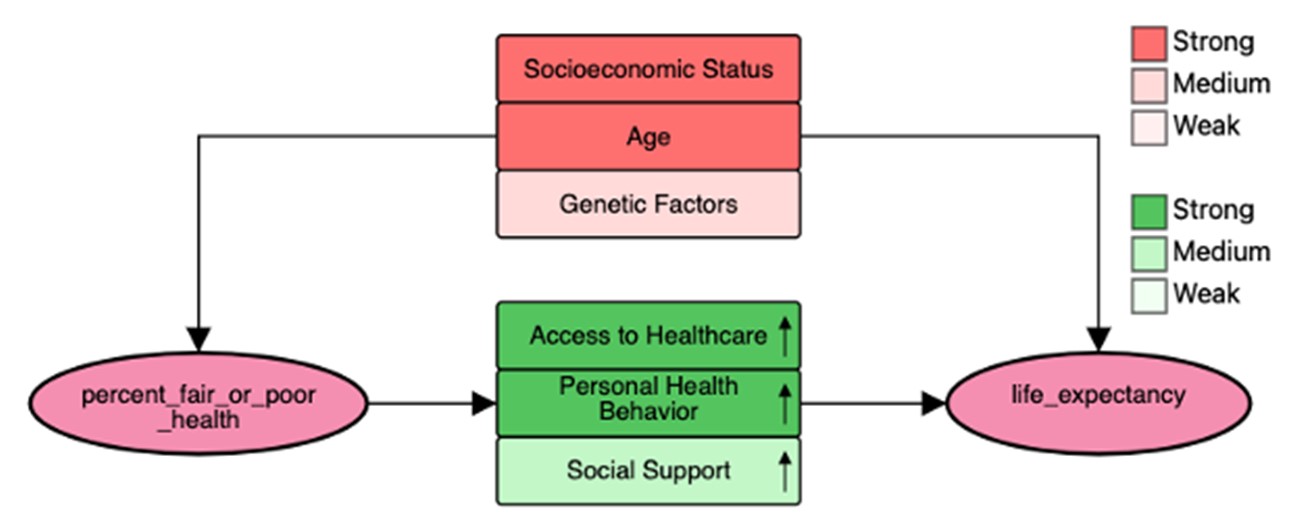}
\caption{Causal Relation Environment Chart for an improbable level combination of the relation \textit{Percent Fair or Poor Health - Life Expectancy}, namely one where both variables have positive levels. The up arrows in the mediators show how this improbable combination might be achieved, in form of interventions on the mediators in the direction of the arrows. }
\vspace{-15 pt}
\label{fig:HPH-HLE}
\end{figure}

Fig. \ref{fig:HPH-HLE} explores the unlikely relationship of \textit{high PFPH} causing \textit{high LE}. GPT-4 correctly identifies this as "counterintuitive" but treats it
as a ’hypothetical scenario’. It suggests that a mediating relationship
would need to exist to achieve the desired \textit{high LE}. These mediators represent potential points of intervention that policymakers could target to increase \textit{high LE}, despite the \textit{high PFPH}. For example, a policymaker might intervene by opening additional health clinics in areas with \textit{high PFPH}, thereby increasing the mediator, \textit{Access to Healthcare}, which in turn improves \textit{high LE}.

\begin{figure}[b!]
\vspace{-15 pt} 
\centering
\includegraphics[width=0.7\linewidth]{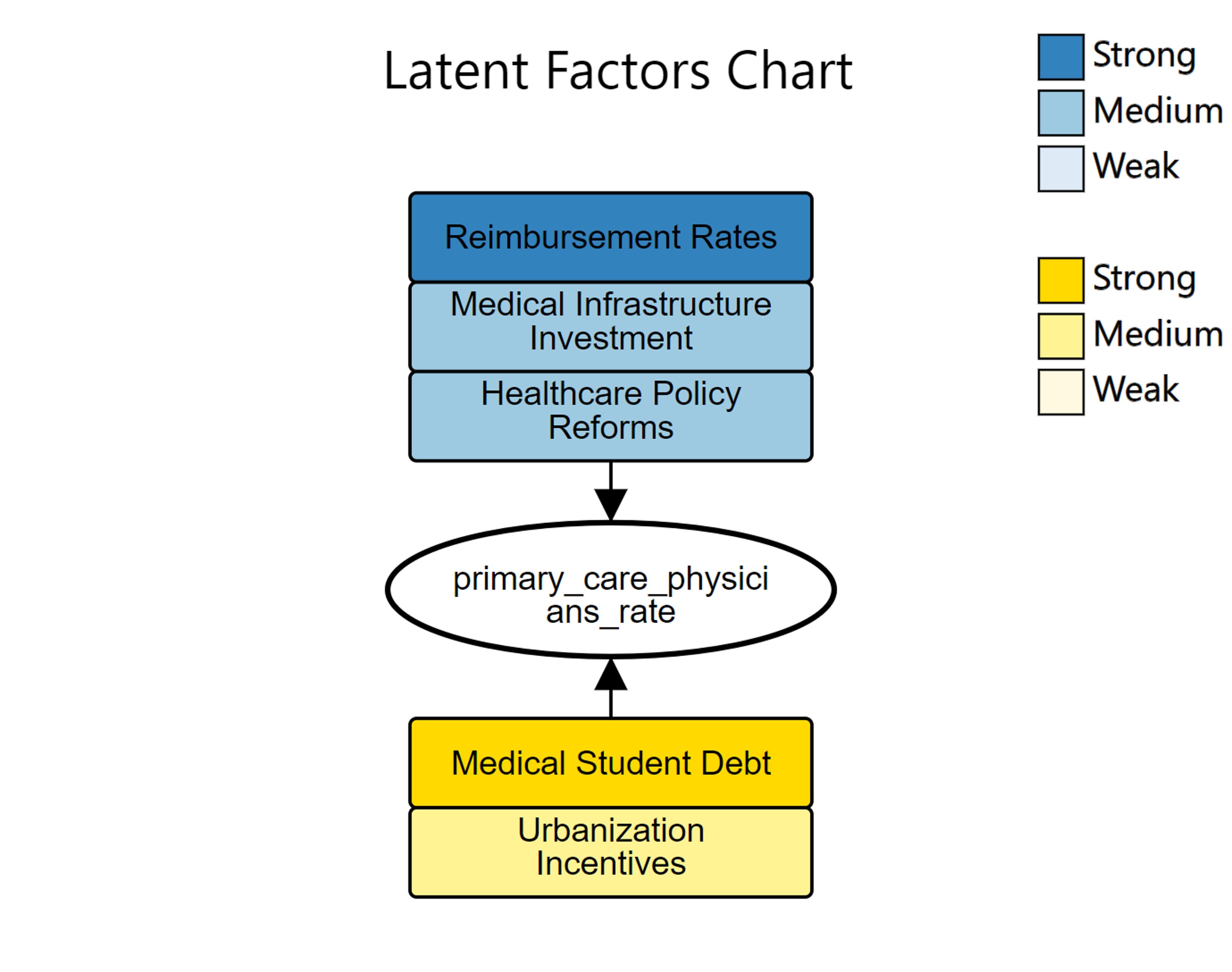}
\vspace{-10 pt}
\caption{Latent Factors Chart for \textit{Primary Care Physicians Rate} as the target variable. The blue nodes above are factors with positive influence, while the yellow nodes below are factors with negative influence, intensity codes strength (see color legend on the right).}
\label{fig:latent-factors}
\vspace{-5 pt} 
\end{figure}

\subsection{The Latent Factors Chart}
The Latent Factors Chart is designed to advise on supplementary variables that should be included in the analysis of the current causal graph. It highlights the variable of interest alongside other potential factors that may directly influence it. These latent factors are color-coded: different shades of yellow represent negative effects, while different shades of blue represent positive effects. By reviewing the provided justifications, users can verify the relevance of these latent factors and incorporate the most significant ones into the causal graph. For example, as shown in Fig. \ref{fig:latent-factors}, \textit{Reimbursement Rates} is identified as a strong positive influence on the \textit{Primary Care Physician Rate}. GPT-4 justifies this by stating, 'Higher reimbursement rates for primary care services can make the field more financially appealing, attracting more physicians to primary care and directly influencing the primary care physician rate.' On the other hand, as discussed in section \ref{propmt_latent}, \textit{Medical Student Debt} is a strong negative factor, with high levels of debt deterring graduates from entering lower-paying specialties like primary care. To improve the primary care physician rate, a policy analyst may update the causal graph to emphasize increasing reimbursement rates and reducing medical student debt as key intervention points.

\subsection{Model Tree}
Although DAGs provide intuitive representations for causal graphs, their scalability is limited, with typical graphs containing an average of 12 nodes, with most graphs having between 9 and 16 nodes \cite{tennant2021use}. To address the challenge of representing causal relationships involving dozens or even hundreds of variables while still using DAGs, we introduce the Model Tree (see  Fig. \ref{fig:teaser}(A)).

The Model Tree is an N-ary tree structure in which each node represents a distinct causal graph. The root node corresponds to the global model, providing an overview of the entire system, while child nodes represent progressively more specific or local models. Users interact with the general model at the root and can select a subset of nodes to create a child node in the Model Tree. These child nodes inherit a subgraph from the parent node, composed of the selected nodes and their associated edges. As users continue to refine and expand the causal graph at each hierarchical level, the Model Tree evolves into a structured hierarchy of causal graphs, offering different levels of granularity and detail.

The Model Tree also supports personalization of causal models. Bidirectional effects, or feedback loops, occur when two variables influence each other over time \cite{murray2022wheel}. A classic example is the relationship between obesity and depression, where each can be a cause of the other over time. While such bidirectional effects are prevalent in fields such as epidemiology, biology, etc., they conflict with the acyclic nature of DAGs. The Model Tree bypasses this issue by splitting a causal model with a bidirectional edge into two causal models with unidirectional edges, thereby maintaining acyclicity. This approach aligns with the concept of personalized causality or causal heterogeneity, where the direction of causal relationships may differ between individuals or for the same individual at different points in time.

\subsection{Graphical Encoding Schemes and Dashboard}

In the graphical encoding schemes, we sought to maintain clarity across different diagrams while ensuring consistent meaning for shapes and colors that represent similar concepts. As mentioned, in the causal diagram, red lines (varying in thickness) represent positive causal relationships, and green lines represent negative causal relationships. The variation in thickness reflects the strength of these relationships. Different shapes provide context: oval-shaped purple marks the outcome variable, cyan denotes a selected node, unfilled ovals signify causal nodes, and dotted shapes/lines indicate elements derived from GPT-4 that are not yet confirmed by data. 

In the Causal Debate Chart, using magenta bars for increasing levels and sky blue bars for decreasing levels provides a clear visual contrast, ensuring that this chart remains visually distinct from the others, especially in its portrayal of variable levels. The Causal Environment Chart uses red shaded boxes for confounders and green shaded boxes for mediators, maintaining consistency with the causal diagram’s color scheme but applying the colors to shaded boxes (instead of lines), which helps differentiate between the two types of charts.

Finally, the Latent Factors Chart introduces blue shaded boxes to represent positive influence and yellow shaded boxes for negative influence. These colors were chosen to avoid overlap with the red and green used in other charts, allowing the latent chart to stand apart while still adhering to a recognizable positive/negative color scheme. 

All charts and GPT-4 justifications are accessible via an interactive dashboard that implements these graphical encodings. Fig. \ref{fig:teaser} shows an example of this dashboard for the first usage scenario, presented next. 

\section{Usage Scenarios}
\label{sec:usage-scenario}

In this section, we demonstrate the capabilities of CausalChat by presenting two usage scenarios that employ real-world datasets.

\textbf{The AutoMPG Dataset}\cite{misc_auto_mpg_9} covers 398 cars from the 1980s. Each car is characterized by 8 attributes: origin, model year, weight, horsepower, displacement, acceleration (time to 60 mph), cylinders, and miles per gallon (mpg). Due to their simple mechanics, 1980s cars exhibit straightforward causal relationships among the variables.

\textbf{The Opioid Death Dataset} combines 9 key socioeconomic factors sourced from the County Health Ranking database \cite{health2023ranking} with opioid death data from the CDC WONDER database \cite{cdc2021opioid} for over 3,000 US counties. The chosen factors are hypothesized to have either direct or indirect impacts on opioid-related deaths.

\begin{figure}[t!]
\centering
\includegraphics[width=\linewidth]
{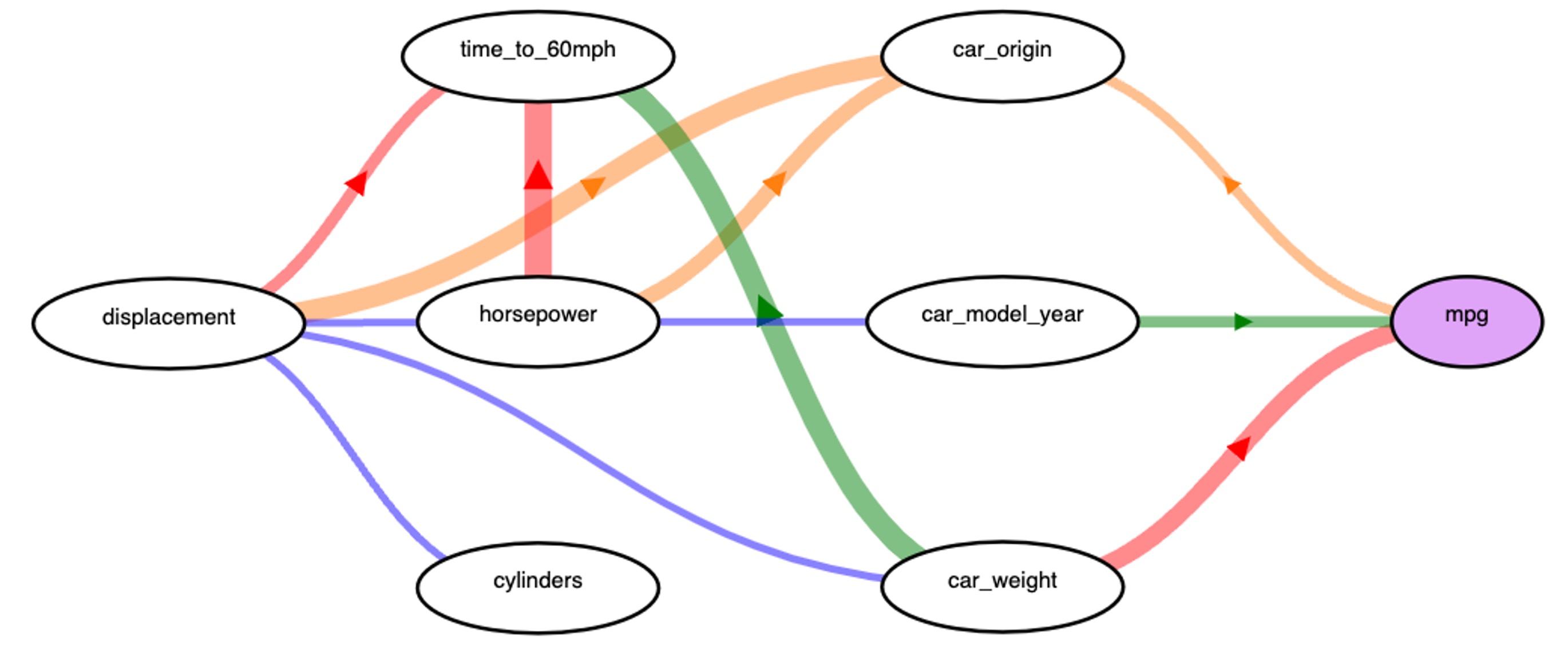}
\vspace{-10pt}
\caption{Initial causal graph of the AutoMPG dataset generated by the GES algorithm. Green (red) directed edges indicate positive (negative) causation, blue undirected edges link variables that are correlated and potentially causal, yellow causal edges connect categorical variables.}
\label{fig:init-graph}
\vspace{-10pt}
\end{figure}

\begin{figure}[t!]
\centering
\includegraphics[width=0.9\linewidth]
{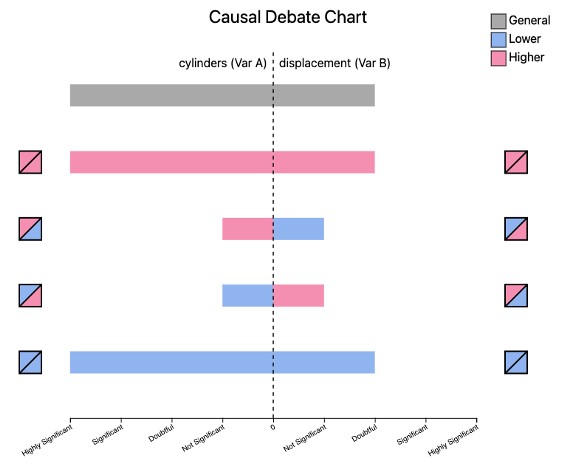}
\vspace{-5pt}
\caption{Causal Debate Chart of the relation \textit{Cylinders} - \textit{Displacement}.}
\label{fig:debate-cl-ds}
\vspace{-15pt}
\end{figure}

\subsection{Exploratory Causal Analysis: Automotive Engineering}

In this study, we focus on Oscar, an automotive hobbyist who wants to gain more insight into automotive engineering. Oscar finds the AutoMPG dataset and reads it into CausalChat. He then employs the GES algorithm and obtains the preliminary causal graph shown in Fig. \ref{fig:init-graph} (see caption for an explanation of the edge coloring). We now follow Oscar in his mission to audit and expand this causal graph. 

\textbf{Resolving undirected edges.} Oscar identifies four blue edges that the GES algorithm could not resolve, possibly due to the dataset's limited coverage of the car domain.  He employs CausalChat's GPT-4 suite to address these edges. For brevity, we shall focus on how he resolves the blue edge between \textit{Cylinders} and \textit{Displacement}, the procedure for the other blue edges is similar.

\begin{figure*}[t!]
  \centering
  \includegraphics[width=\linewidth]{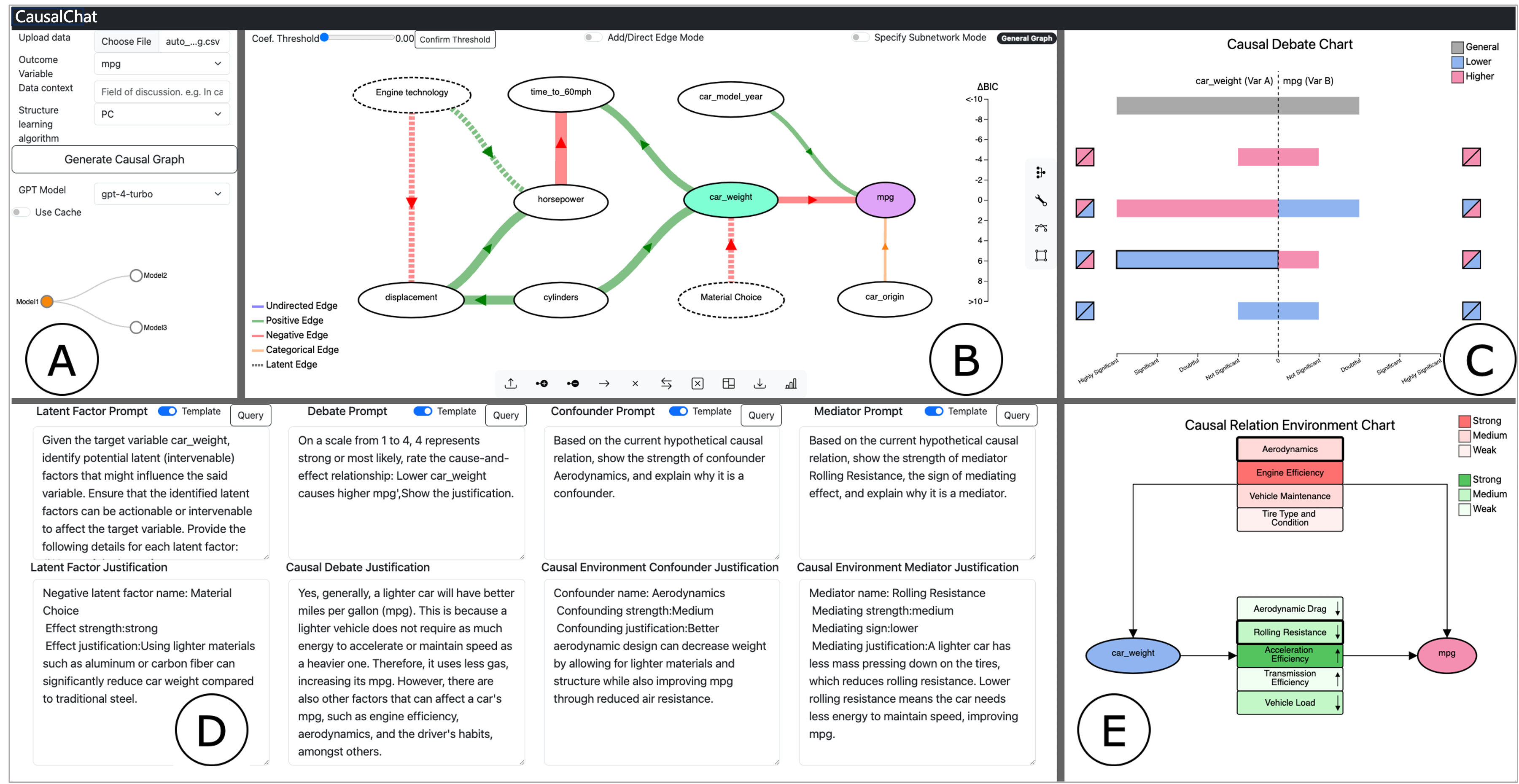}
  \caption{
The CausalChat Dashboard analyzing the AutoMPG dataset. It includes (A) the Control Panel for specifying fundamental parameters and tracking the model tree for version control, (B) the Causal Graph Panel for interactive refinement of the graphical model, (C) the Causal Debate Chart for resolving causal directions and inclinations, (D) the Causal Justification Panel offering a rationale for each hypothetical causal statement, covering latent factors, potential confounders, and mediators, and (E) the Causal Relation Environment Chart suggesting potential latent variables, confounders, and mediators for specific causal relations and variables. In this figure the confounder/mediator chart is shown.
  }
  \label{fig:teaser}
  \vspace{-10pt}
\end{figure*}

 Oscar clicks on the blue edge and the system generates the Causal Debate Chart depicted in Fig. \ref{fig:debate-cl-ds}. It is immediately apparent that the bars representing \textit{Cylinders} are notably longer than those for \textit{Displacement}, for the general grey bars as well as for the red-red and the blue-blue bars and all at full strength -- a classic pattern for a positive causation. Oscar contemplates converting the blue edge to a green directed link from \textit{Cylinders} to \textit{Displacement}. After confirming this direction from the GPT-4 justification panel (see an example in Fig. \ref{fig:teaser}D),  
 Oscar directs the edge as suggested. In a similar fashion he also directs the other blue undirected edges. 



\textbf{Adding confounders and latent variables.} Oscar is still curious about the relation of \textit{Displacement} and \textit{Horsepower}. He examines the corresponding Causal Relation Environment Chart (not shown) and identifies \textit{Engine Technology} as a confounder. Furthermore, he also discovers a latent factor -- \textit{Material Choice} -- which exerts a reducing effect on \textit{Car Weight}. All of these interactions taken together give rise to the final causal graph shown in Fig. \ref{fig:teaser}B. The edges for the two newly added variables are visualized as dotted lines to convey that their weights have not been calculated from data yet -- all Oscar has are the strengths indicated by GPT-4.

Inspecting the triad \textit{Displacement},  \textit{Horsepower}, and \textit{Engine Technology}  reveals that \textit{Horsepower} is, in fact, a causal collider.  It is influenced by both \textit{Displacement} and \textit{Engine Technology}, or dominated by one of the two. While in older cars displacement was the dominant factor determining horsepower, modern engine technology has altered this role. The negative causal effect of \textit{Engine Technology} on \textit{Displacement} further suggests that as engine technology advances, the need for displacement to elevate horsepower diminishes. In other words, higher values of engine technology have become more influential in determining horsepower than displacement alone. This interpretation aligns with the idea that modern engines, with advancements in technology, can achieve higher horsepower despite low displacement. In essence, by adding \textit{Engine Technology}, Oscar has modernized the original causal network derived from the antiquated dataset of 1980s cars. 


\textbf{Adding mediators.} In the updated causal graph Oscar notices the antagonistic relationship of \textit{Car Weight} and \textit{Horsepower} which affect \textit{Time to 60 MPH} in opposite ways. Having resolved the need for high \textit{Displacement} also reduces the need for a high number of \textit{Cylinders} which would cause high \textit{Car Weight}. However, there may be other valid causes for high \textit{Car Weight} not represented in the graph. Consequently, Oscar chooses to investigate potential ways to mitigate these factors.

\begin{figure}[t]
\centering
\includegraphics[width=\linewidth]
{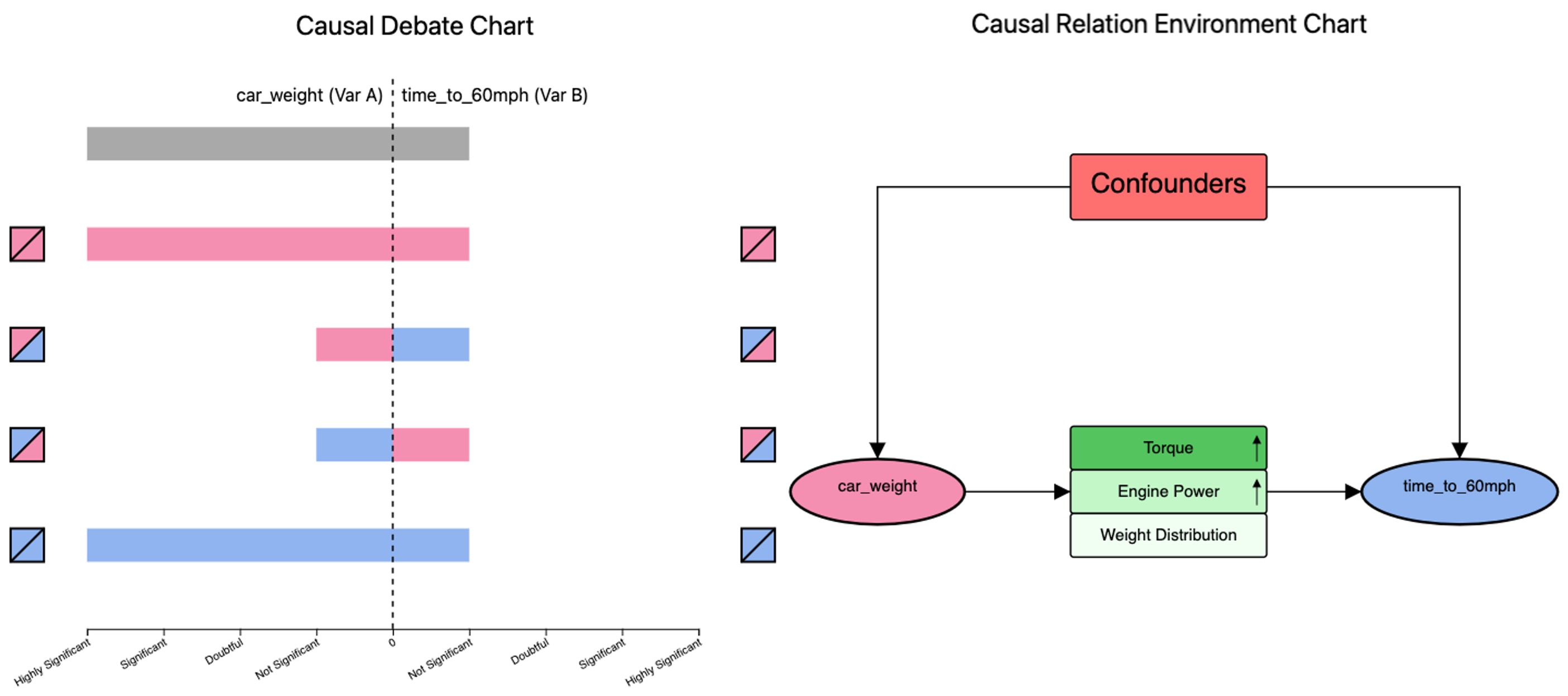}
\vspace{-15pt}
\caption{Assessing the hypothetical causal relation where a low time to 60 MPH could be achieved despite high car weight.}
\label{fig:weight-time}

\vspace{10 pt}

\centering
\includegraphics[width=\linewidth]
{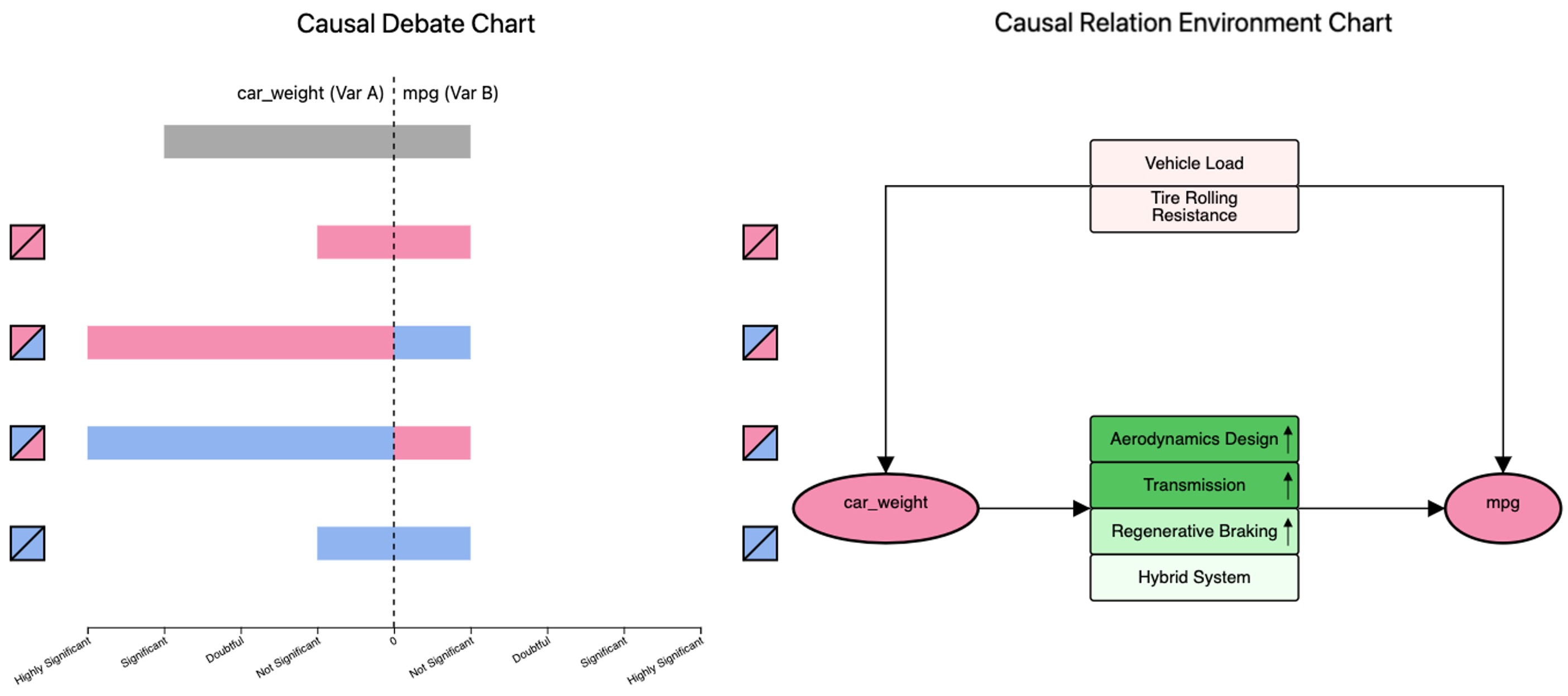}
\vspace{-15pt}
\caption{Assessing the hypothetical causal relation where high MPG can be achieved despite high car weight.}
\label{fig:weight-mpg}

\vspace{10 pt}

\centering
\includegraphics[width=\linewidth]
{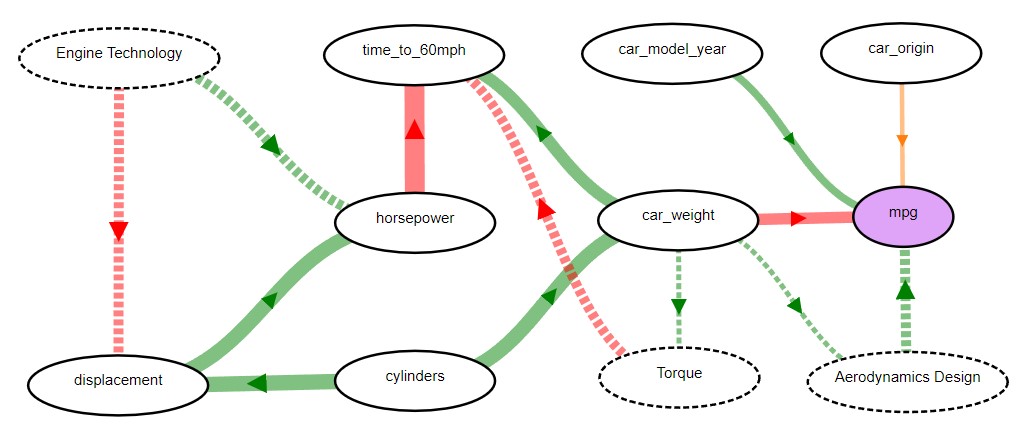}
\vspace{-15pt}
\caption{Causal graph that incorporates all of Oscar's innovations.}
\label{fig:innov-graph}
\vspace{-15 pt}
\end{figure}

The Causal Debate Chart in Fig. \ref{fig:weight-time} (left) illustrates that heavy cars take more time to reach 60 MPH (2\textsuperscript{nd} bar pair) than lighter cars (5\textsuperscript{th} bar pair), i.e., they have poor acceleration. These are the hard facts represented by the two long bars. But Oscar wants to innovate and is looking for a car that can be heavy yet quick off the mark, the condition represented by the 
3\textsuperscript{rd} pair of bars.
Clicking on the left bar brings up the corresponding Causal Relation Environment Chart,  Fig. \ref{fig:weight-time} (right). It suggests that increasing the engine's twisting power, or torque, can improve acceleration and mitigate the effect of high car weight. 

This prompts Oscar to include \textit{Torque} as a mediator between \textit{Car Weight} and \textit{Time to 60 MPH}, 
see Fig. \ref{fig:innov-graph}. It introduces \textit{Torque} as an additional causal factor that opposes the impact of high \textit{Car Weight}. This influence is indicated by the red dotted outgoing edge and its purpose is indicated by the green dotted incoming edge, i.e., heavier cars need more torque. 





To further optimize his ideal car, Oscar shifts his attention towards enhancing its mileage efficiency. The causal graph reveals an inverse relation link between \textit{Car Weight} and \textit{MPG}. Using a similar process as above he refers to the Causal Debate and Causal Relation Environment Charts in Fig. \ref{fig:weight-mpg}, now for the case of high \textit{car Weight} and high \textit{MPG}. He learns that incorporating advanced \textit{Aerodynamic Design} principles in the car's design can markedly cut down fuel consumption during operation, ultimately leading to a notable improvement in \textit{MPG}. It advises car designers to apply aerodynamic principles when a car is heavy, as indicated by the green causal edge.

Oscar also inserts this mediator into the \textit{Car Weight} and \textit{MPG} relation, yielding the final causal graph in Fig. \ref{fig:innov-graph}. 



\subsection{Causal  Strategizing for Opioid Mortality Prevention}
Here, we join Lena, an epidemiologist, on her mission to discover preventive measures against the widespread opioid epidemic afflicting numerous counties in the United States. 
She starts out with 9 socioeconomic variables that she feels are related to opioid mortality plus data on opioid mortality itself
(the aforementioned opioid death dataset). 

\begin{figure}[h]
\centering
\includegraphics[width=\linewidth]
{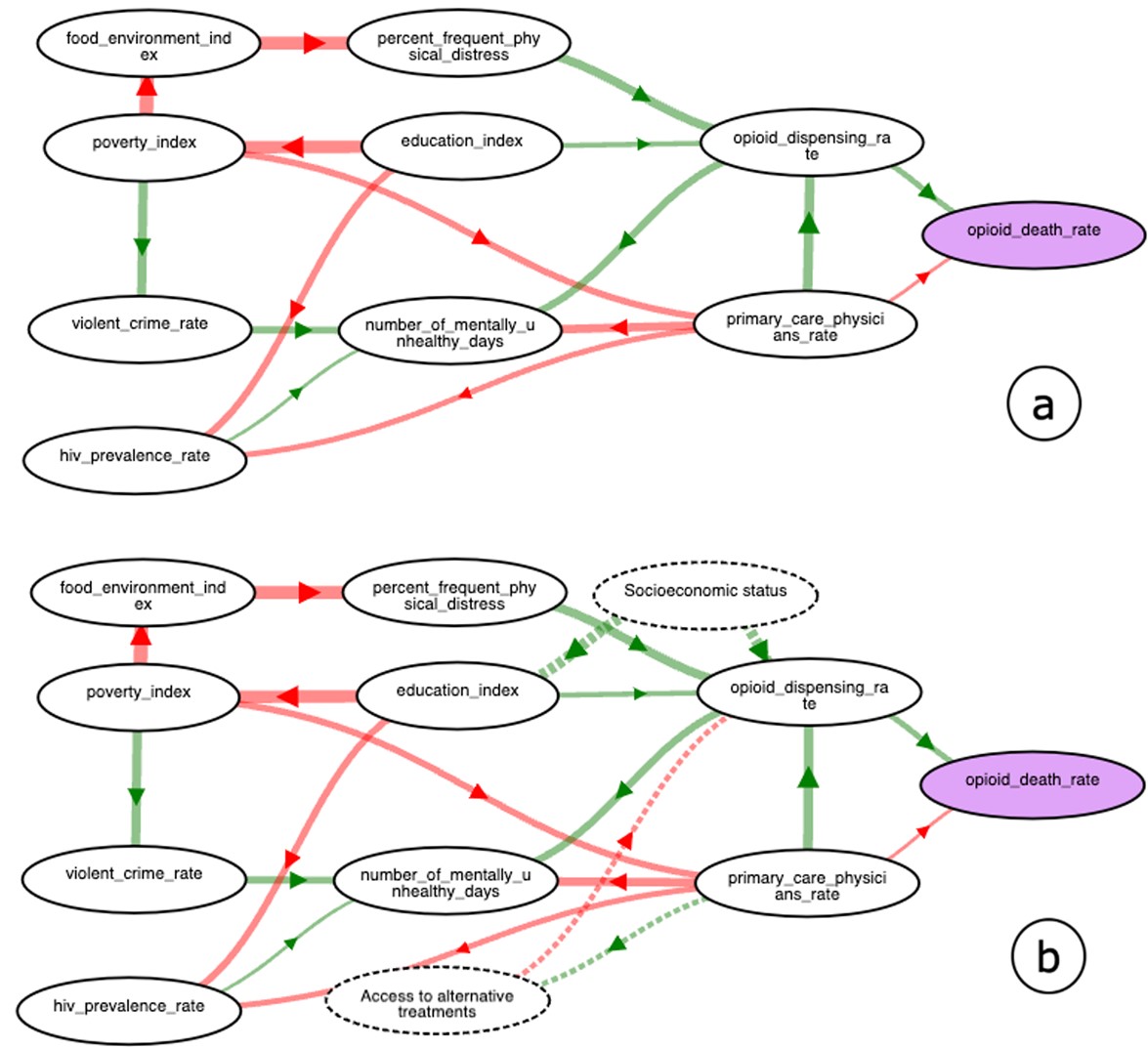}

\caption{Causal graphs created by Lena: (a) Initial model leveraging domain knowledge. (b) Model enhanced via CausalChat assistance, incorporating confounders and mediators for actionable recommendations.  }
\label{fig:usg2_causal_graphs}
\vspace{-15pt}
\end{figure}

\begin{figure}[]
\centering
\includegraphics[width=\linewidth]
{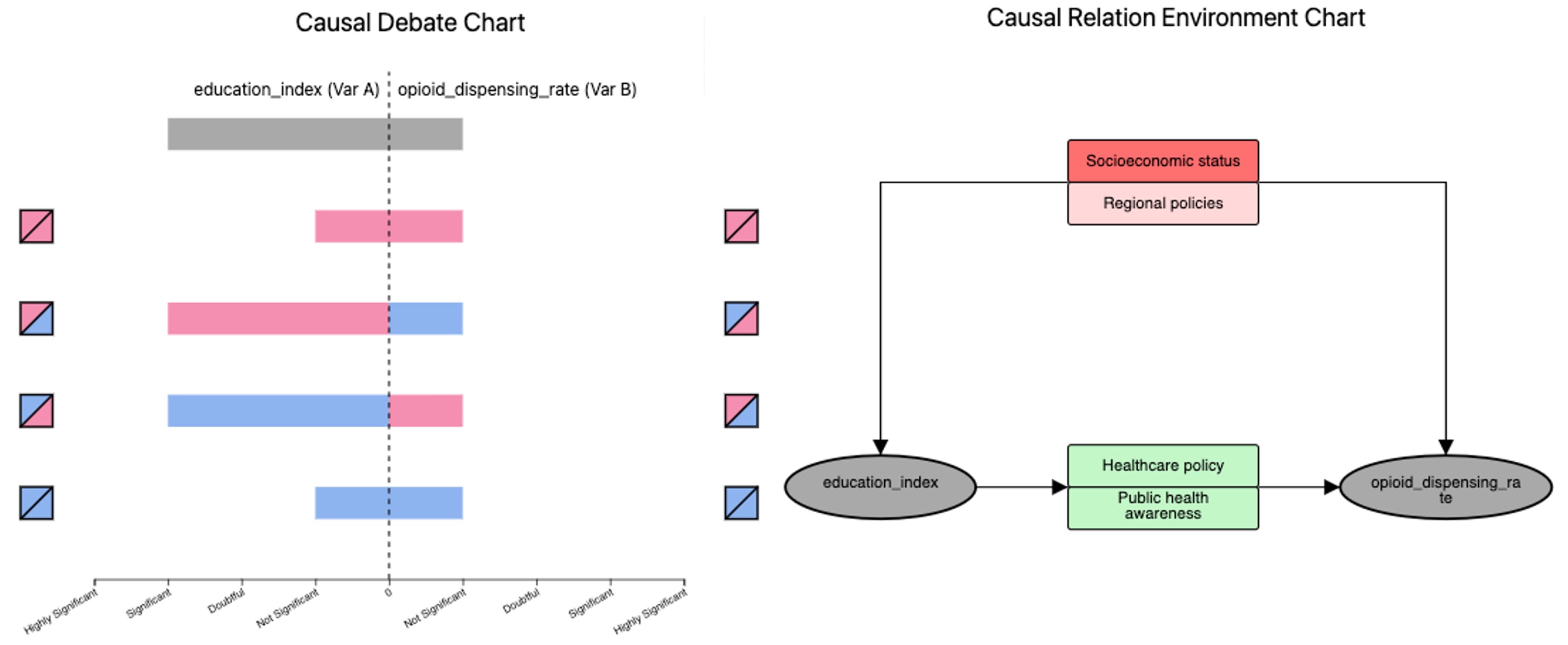}
\vspace{-15pt}
\caption{The weak direct causal relations (left) suggest a confounding between \textit{Education Index} and \textit{Opioid Dispensing Rate} (right).}
\label{fig:usg2_dbt_env_edu}

\bigskip

\centering
\includegraphics[width=\linewidth]
{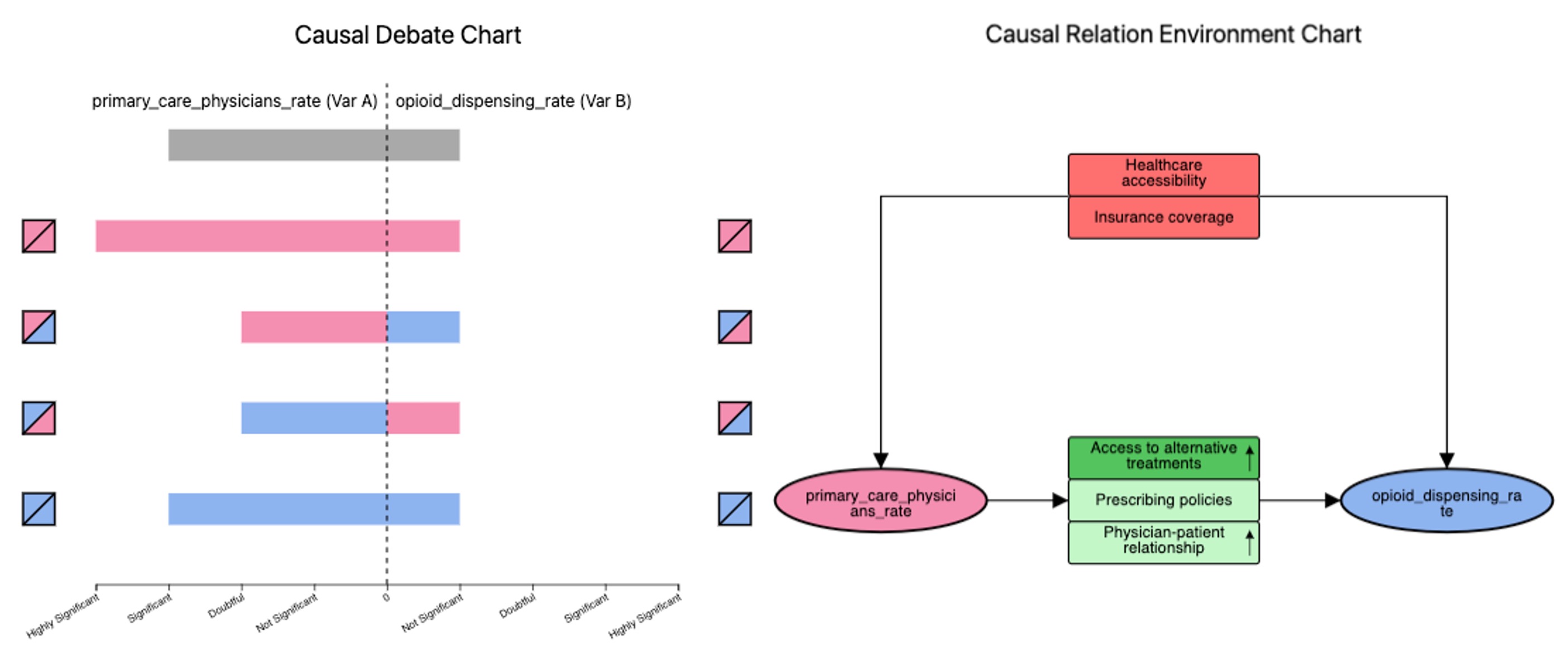}
\vspace{-15pt}
\caption{Gaining control over opioid dispensing through mediators.}
\label{fig:usg2_dbt_env_primary}

\bigskip

\centering
\includegraphics[width=\linewidth]
{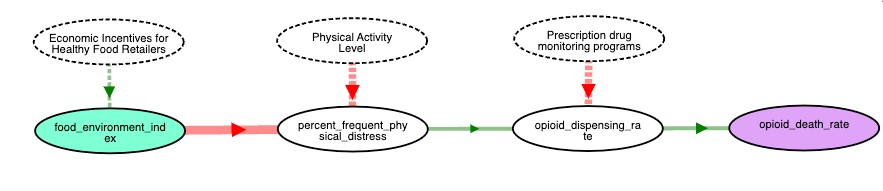}
\vspace{-15pt}
\caption{A causal pathway delineating multiple intervenable latent factors along a critical causal chain for a specific population group.}
\label{fig:usg2_final_chain}
\vspace{-15 pt}
\end{figure}

\textbf{Initial setup.} As a first step, Lena utilizes her expertise to construct a foundational causal graph based on the available data, see Fig. \ref{fig:usg2_causal_graphs}a. However, she finds herself dissatisfied with the connection between \textit{Education Index} and \textit{Opioid Dispensing Rate}. Intuitively, she believes that enhanced education would raise awareness of the adverse effects associated with opioid dispensing. Yet, the green edge she initially drew suggests the opposite. To investigate, she opts to reassess this edge using the Causal Debate Chart shown in Fig. \ref{fig:usg2_dbt_env_edu} (left).

\textbf{Exploring doubts.} Examining this chart, Lena comes to realize that while there is a slight inclination towards the current causality, none of the bars exhibit highly significant strength, indicating a raised likelihood of a confounder. She clicks on the left bar of the general (grey-colored) relation which brings up its Causal Relation Environment Chart. Indeed, two confounders are suggested, with \textit{Socioeconomic Status} being the strongest. The  justification states that "socioeconomic status can influence level of education and also lead to better access to healthcare facilities where more prescription opioids are dispensed". This confirms Lena's initial apprehension  about the direct edge, with the justification pointing to an important mediator between \textit{Socioeconomic Status} and \textit{Opioid Dispensing Rate}: \textit{Prescription}. Thus, a mitigating policy intervention would be to tighten opioid prescription regulations.


\textbf{Addressing the target effect.} Next, Lena sets out to identify actionable measures to help reduce opioid fatalities. She directs her attention to the edge from \textit{Primary Care Physician Rate} to \textit{Opioid Dispensing Rate}. Generally, the relationship is positive since opioid dispensing typically involves doctors\footnote{Here we consider the original source of opioids: physicians. However, in modern times, opioids often stem from the illicit distribution of fentanyl.}  as is indicated by the dominant 2\textsuperscript{nd} bar pair in the associated Causal Debate Chart, Fig. \ref{fig:usg2_dbt_env_primary}. In search of an intervention, Lena focuses on the hypothetical, but more desirable relationship just below this pair: higher \textit{Primary Care Physician Rate} leading to lower \textit{Opioid Dispensing Rate}. The associated Causal Relation Environment Chart offers numerous valid mediators, such as \textit{Access to Alternative Treatments}. This suggests that if alternative non-opioid treatments are made available, the opioid dispensing rate and its subsequent use can be reduced. Fig. \ref{fig:usg2_causal_graphs}b shows the updated causal graph.


\textbf{Focusing on a specific population group.} Lena now continues her exploration with a focus on a specific causal pathway: \textit{Food Environment Index} → \textit{Percent Frequent Physical Distress} → \textit{Opioid Dispensing Rate} → \textit{Opioid Death Rate}. This pathway tells a compelling story: individuals experiencing a food environment crisis may suffer physical distress, leading them to rely on opioid-containing painkillers and, ultimately, succumb to addiction and death. Lena is determined to aid this vulnerable population, utilizing CausalChat's Create Sub-Graph module to isolate these variables and associated edges, creating a new child node in the model tree. Her objective now is to identify efficient direct factors that can influence this causal pathway. Using Latent Factor Charts (not shown), Lena identifies three intervenable measures to improve this pathway: economic incentives for healthy food retailers, physical activity promotion, and prescription drug monitoring programs (see Fig. \ref{fig:usg2_final_chain}). However, despite the rationale behind these latent factors, their effectiveness in reducing opioid death rates cannot be conclusively determined and would require research with real data and experimentation. At this juncture, they are merely conceptualized ideas facilitated by Lena's utilization of CausalChat. 



\section{User Study}
We conducted a two-fold user study: (1) Obtaining feedback from domain experts to assess the practical applicability and logical coherence of our proposed framework; (2) Evaluating the usability and efficacy of our framework by assigning tasks to non-expert users.

\subsection{Datasets}
\label{sec:datasets}
In these user studies, we utilized two real-world datasets related to public health. Each data point corresponds to a distinct county in the United States for the year 2019. The expert assessment utilized the opioid death dataset described in Fig. \ref{sec:usage-scenario}. The non-expert user study employed the Life Expectancy dataset detailed below.

\textbf{The Life Expectancy dataset} comprises 8 key variables sourced from the County Health Rankings \& Roadmaps Database \cite{health2023ranking}:  firearm fatality
rate, violent crime rate, average grade performance, high school graduation rate, food environment index, percent fair or poor health, primary
care physician rate and debt income ratio for each of more than 3,000 US counties, All of these variables are
recognized to affect demographic life expectancy either directly or indirectly.


\subsection{Expert Assessment}
We invited three domain experts, affiliated with our university, to evaluate CausalChat. These experts specialize in health policy (P1), exposure science (P2), and environmental epidemiology (P3). All possess solid background in causal inference and have a shared interest in utilizing quantitative approaches in public health modeling.

Each of the three sessions took place via Zoom. We first introduced our interface, emphasizing the functions and significance of each visual component. We ensured that the experts were familiar with how to navigate the system. Subsequently, they were tasked with refining a causal graph containing two unresolved edges and one misdirected edge, with the goal to achieve a valid causal graph, and eventually ideating additional factors. All were able to achieve these tasks. Throughout their interaction with the system, we encouraged them to articulate their thought process. 
Finally, we gathered their feedback on CausalChat, specifically soliciting suggestions for potential enhancements. In the following we grouped the verbal session outcomes into specific themes.  

\textbf{Overall assessment.} 
All three participants unanimously praised CausalChat as an exceptional ideation tool, noting its effectiveness in helping non-experts quickly grasp the most prominent causal relations between variables in a field potentially unfamiliar to them. Specifically, P2 highlighted the efficiency of the tool, noting, "CausalChat enlightens non-expert users in a productive fashion. Users don't have to go through an exhaustive literature research process to obtain a fundamental understanding of a new field."  

\textbf{Validation.} P1 expressed that her expertise led her to anticipate the presence of \textit{Withdrawal Treatment} as a mediator between opioid dispensing  and opioid-related deaths. This anticipation was validated when she discovered \textit{Withdrawal Treatment} listed as a significant mediator within the Causal Relationship Environment chart. 
She added that once identified conceptually, the proposed mediators, confounders, and other new variables can be statistically tested,
enhancing confidence and adding accuracy beyond GPT-4's strength assessments. 

\textbf{Making access to domain science easier and more streamlined.} The potential of our tool to provide expedited access to domain knowledge became evident when the experts confirmed CausalChat's adeptness in identifying confounders and mediators within an existing causal graph. P1 remarked that "compared to the traditional method of studying and including all possible confounders, which demands relentless and tedious literature review, CausalChat offers a far more efficient solution by automatically and visually presenting all potential confounders. This feature significantly enhances accessibility to science".

\textbf{Use as a research tool.} P3 perceived CausalChat as an actual research planner. She said: "I would use the framework for planning my research. With all these essential variables and their information visually represented, I can have a clear goal of what data I need to collect, what research papers are useful, and what academic areas I can potentially contribute to." She further noted that the synergy between the Causal Debate Chart and the Causal Relation Environment Chart not only highlighted well-studied areas but also suggested intervention strategies (such as mediators) and ways to mitigate unrealistic relationships.  Relatedly, P1 remarked that the framework effectively prompts domain experts to consider essential components that may be overlooked but are crucial for enhancing the outcome variable.





 
\textbf{Preferred alternative to pure data-driven causal analysis.} P2, an expert in experimental-based causality, began his session by voicing strong skepticism with regards to existing methods for automated causal analysis. He contended that observational data could only uncover correlations, not causation. According to his view external information is indispensable for directing causal edges, and even for forming them in the first place. He favors the alternative approach supported by CausalChat -- taken by Lena in our second case study -- which first constructs and refines a causal graph by leveraging the knowledge of GPT-4 and then estimates the causal effect of each edge using data.

\textbf{Satisfying the need for personalized causal models.} P2 emphasized that socioeconomic variables are predominantly cross-sectional and can engender feedback loops, resulting in bidirectional edges. This poses a challenge to the assumption of the causal graph being a DAG, underscoring the importance of further expanding our model tree panel. In each distinct causal model, there exists a distinct data generating process where variables have a clear upstream and downstream direction. An edge pointing from A to B in one model can point reversely in another. These two models will share a parent model in our model tree. After learning about this feature, P2 acknowledged its effectiveness in addressing his concern.

\textbf{Concerns with GPT-4's omission of citations.} Both P1 and P3 expressed concerns about the inherent uncertainties of GPT-4. On one hand, CausalChat lacks the ability to provide citations when clarifying a standpoint or rating a causal relation, and the mechanism by which GPT-4 analyzes papers for causation lacks transparency. On the other hand, there is a risk that, given the varying quality of existing literature, GPT-4 may not be able to differentiate between rigorous and non-rigorous research. While this may not pose a problem for well-trained scientists who can filter out moderate papers, there is concern that users might rely on CausalChat for decision-making and consider it the ultimate truth. P1 suggested including a disclaimer indicating that some edges may be misdirected or omitted.

\subsection{Non-Expert User Study}
\label{sec:nonexpertuserstudy}
To understand how CausalChat can help laypeople explore unfamiliar areas of knowledge, we conducted an ablation study to contrast and validate the following three hypotheses:

\begin{enumerate}[label=\textbf{H\arabic*.}, leftmargin=*, labelsep=1em]
    \item  CausalChat provides comprehensive yet concise guidance to help users uncover interrelationships among variables. Additionally, CausalChat encourages users to identify latent third variables, such as confounders and mediators, that may influence the relationship between two variables.
        
    \item CausalChat enhances users' efficiency in acquiring knowledge in unfamiliar fields, particularly by improving their ability to uncover interrelationships and identify latent variables, as in H1.
    \item CausalChat is user-friendly, convenient, and effective. 
\end{enumerate}

To test these hypotheses, we conducted an ablation study with three layers. 
The first layer relies on traditional causal model reasoning using the BIC score as a quality metric, the second layer incorporates text prompts for an LLM, and the third layer integrates these methods into the CausalChat interface. Each layer progressively adds more sophisticated tools to enhance causal reasoning and decision-making.



\begin{enumerate}[label=\textbf{L\arabic*.}, leftmargin=*, labelsep=1em]
  \item \textbf{Conventional BIC-Score Based Causal Reasoning:} In this layer, users refine causal graphs by combining their domain knowledge with feedback on modifications, evaluated through the BIC score.
  
   \item \textbf{LLM-support with text:} This layer builds on L1 by incorporating a ChatGPT interface. After specifying the relationship of interest, users receive exemplar prompts for text responses to explore causal connections, allowing them to determine the most plausible causal relation based on multiple GPT-4 responses addressing various aspects of the relationship.
   
    \item \textbf{CausalChat Lite:} 
    This layer adds to L2 some of the visual elements of CausalChat: (1) the causal debate charts, (2) the causal relation environment charts, and (3) the justification panel. 
\end{enumerate}

Our study focuses on CausalChat's ability to correct distorted effects, identify mediation effects for indirect relationships, and resolve edge directionality in causal graphs. The aim is not to encourage users to endlessly expand the graph but to guide them in accurately determining the direction of undirected edges and addressing omitted variables that are critical within the scope of the theme under consideration.

\textbf{Dataset.} To minimize variation in the final graph, we used a subset of the Life Expectancy dataset. This subset includes the following variables: food environment index, percent in fair or poor health, primary care physician rate, debt-to-income ratio, and life expectancy.

\textbf{Participants.}
We recruited six university students (3 males, 3 females) for usability studies. All participants were familiar with web browser-based frameworks and could participate either in person or via Zoom. None of the participants had expertise in epidemiology.

\textbf{Study Design.}
Participants were guided through the study using Qualtrics, an online survey tool that follows a predefined workflow. The workflow began with tutorials on data causality and instructions for using CausalChat, followed by a quiz to reinforce participants' understanding of graphical causal models, including terms related to confounders and mediators. Participants had to answer all quiz questions correctly before proceeding to the main tasks involving CausalChat evaluation. Finally, they completed a series of usability questions, including the standardized System Usability Scale (SUS) questionnaire, to provide feedback on their experience.

\textbf{Ablation Study Stages.} We divided a session into three stages, each corresponding to the testing of a different layer (method). At each stage, participants were asked to resolve undirected edges using the tools available for that stage. For each undirected edge, participants had the option to direct the edge, remove it, add a confounder, or add a mediator. Initially, participants were presented with 9 preset undirected edges derived from the subset of variables in the life expectancy dataset.
To ensure fairness, participants could resolve any 3 undirected edges of their choice at each stage until they were satisfied. They were not allowed to modify any edges they had already edited in previous stages.

\textbf{Stage 1.} In the first stage, participants were asked to resolve 3 undirected edges using BIC score feedback. A bar displayed the change in the BIC score for the affected nodes after each edge modification, enabling participants to assess the impact of their changes.

\textbf{Stage 2.} In the second stage, participants were asked to resolve another 3 undirected edges of their choice using LLM support provided through a set of pre-formulated prompts, similar to those used by CausalChat and consisting of debate prompts and causal relation environment prompts. Participants received text-based responses from GPT-4 based on these prompts and made their decisions by reviewing the responses.

\textbf{Stage 3.} In the final stage, all operations were conducted within CausalChat Lite. For the last 3 edges, participants made decisions based on the visual charts and verified their insights by reviewing the corresponding justifications.

\vspace{5pt}

\noindent \textbf{H1:  Understanding Interrelationships Among Variables}


\noindent \textbf{Stage 1.} In the first stage, participants based their decisions largely on their own knowledge and the BIC score. However, without domain expertise, their decisions were often uncertain or biased. Although the BIC score provided insights into key predictors of the target variable, it was vulnerable to distortions from hidden confounders, which led to inaccuracies in identifying true causal effects. As a result, 28\% (5/18) of the undirected edges were incorrectly resolved. Participants also identified only one mediator and no confounders. When asked why they struggled with latent variables, participants pointed to two main challenges: they tended to focus on relationships they felt most confident about, which were often direct or unrelated, and their lack of expertise made it difficult to recognize potential confounders and mediators, limiting their ability to discover these variables effectively.


\textbf{Stage 2.} In the second stage, introducing GPT-4 as a proxy for domain expertise helped participants address knowledge gaps from stage one. However, participants' performance declined slightly, with 39\% (7/18) of causal edges resolved incorrectly, compared to 5/18 in the BIC score method. This is likely because they tackled the easier edges first in stage one. Nevertheless, we observed that the volume of GPT-4 responses proved somewhat overwhelming, making it difficult for participants to fully comprehend and synthesize the information, despite GPT-4's solid understanding of causal terminology, minimal hallucinations, and useful references to relevant literature. We observed that this guidance was particularly helpful in determining causal directionality and encouraging a deeper exploration of potential confounders and mediators. Although participants struggled at times to process the information, they identified four valid confounders and six valid mediators, with none deemed incorrect. While latent variables remained difficult to identify, the overall decision quality showed a clear improvement compared to stage one.


\textbf{Stage 3.} CausalChat Lite performed the best overall, with only 1/18 incorrect edges (5\%), even though these were the most difficult relations since this stage came last. Participants found it significantly easier to navigate the tasks.
For example, P2 noted: "It becomes much easier after realizing how straightforward it is when the patterns are visualized in the causal debate chart. Before, it was really hard to memorize all the queries at once." The chart simplified the process by visually representing complex relationships, reducing the cognitive load of managing multiple queries simultaneously.

We observed that participants made good use of the debate justification text box to obtain clear reasoning for each causal hypothesis rating.
The causal relation environment chart further supported participants by visualizing confounders and mediators together, offering quick access to potential third variables relevant to their analysis. Across the study, participants identified seven valid confounders and five valid mediators, with no false positives, underscoring the effectiveness of CausalChat in improving the accuracy of causal analysis.

\vspace{5pt}

\noindent\textbf{H2: Efficiency in Causal Auditing Tasks} 

\noindent We assessed the efficiency of CausalChat by measuring two key aspects of the causal auditing tasks: (1) edge modification and (2) the discovery of confounders and mediators. As discussed in the analysis of \textbf{H1}, BIC score based analysis alone is insufficient for guiding users to construct accurate causal graphs. Therefore, our focus here is on comparing the causal auditing efficiency between the LLM-assisted interactions and CausalChat Lite.

For edge editing, we measured the time from when a participant began querying the relationship between two variables until they verbally confirmed satisfaction with the modification. With the LLM support, the length and complexity of responses made reading time-consuming and, for some, tedious. As a result, participants spent an average of 3.3 minutes (SD = 2.6) using the LLM text, while the time decreased to 0.9 minutes on average (SD = 0.9) with CausalChat Lite.

Discovering confounders and mediators involved three steps: (1) querying potential confounders and mediators for a variable pair, (2) evaluating the logical consistency of the identified variables, and (3) adding valid variables to the causal graph. On average, for the LLM-supported version, participants spent 3.5 minutes (SD = 0.5) to add a confounder and 3.6 minutes (SD = 2.1) to add a mediator. Participants spent significantly less time for this task with CausalChat Lite. On average, it took 1.6 minutes (SD = 1.0) to add a confounder and 1.3 minutes (SD = 0.4) to add a mediator. This suggests that CausalChat not only improves efficiency in edge editing but also streamlines the process of identifying and incorporating latent variables.

\vspace{5pt}

\noindent\textbf{H3: Easy to Use} 

\noindent Participants unanimously agreed that CausalChat Lite was the most convenient (M = 4.8, SD = 0.4) and instilled the most confidence (M = 4.2, SD = 0.7) compared to the BIC Score based analysis (BIC) and the LLM text support (LLM). The average confidence score for BIC was 2.5 (SD = 0.8), and 3.3 (SD = 0.7) for LLM. In terms of convenience, BIC scored 2.7 (SD = 1.1), while LLM scored 3.3 (SD = 1.2). CausalChat Lite also achieved a usability score of 79.17 (SD = 6.40) on the standardized SUS questionnaire, placing it in the 85th to 89th percentile—well above the average SUS score of 68 and close to the top 10\% of all SUS scores (80.8).

These results aligned with our expectations for CausalChat in assisting users with causal relation auditing, causal graph refinement, and decision-making. In post-study interviews, participants praised CausalChat as a comprehensive tool for quickly learning about unfamiliar fields. P2 remarked, 'CausalChat did an efficient job condensing the wordiness of GPT-4 responses into more digestible visual infographics, making decision-making easier.' Participants also appreciated the visual designs of the causal debate chart and the causal relation environment chart, noting that once understood, learning causal relationships became much smoother. P4 stated, 'The visualizations are very helpful, especially with the color coding to reflect the direction of mediators or confounding variables.' 

\section{Discussion and Conclusions}
We presented CausalChat, a system that leverages the causal knowledge of large language models (LLMs), particularly GPT-4, to make advanced causal insights accessible to those innovating on complex systems. Our approach addresses skepticism about automated causal inference, which often stems from the challenge of acquiring datasets comprehensive enough to model such systems. Instead of relying on large datasets, we directly query LLMs using carefully crafted prompts. User studies, including those involving participants skeptical of automated causal inference, showed that our LLM-powered tool effectively meets their needs.

While LLMs are not without flaws, CausalChat integrates a robust suite of textual and visual features, allowing users to understand the reasoning behind the causal knowledge provided by the model. These features are essential to the system, bridging the gap between AI-driven insights and human understanding. Our case studies highlight CausalChat's significant potential to enhance users' ability to innovate, improve causal modeling, and advance the study of complex systems across various fields.

Future work will focus on enhancing the scalability of CausalChat's visual interface to accommodate the creation of large causal graphs representing highly complex causal models, most likely via a level of detail approach. Additionally, we aim to develop a crawling mechanism capable of automatically retrieving data for newly added variables and causal relations from online repositories.

\bibliographystyle{IEEEtran}
\bibliography{template}

\vspace{-30pt}

\begin{IEEEbiographynophoto}
{Yanming Zhang}  is currently a PhD student in Computer Science at Stony Brook University. He obtained a Bachelor degree in Mathematics and Applied Mathematics from Southwest Jiaotong University. His research focuses on explainable AI, visual analytics, causal inference, and human-computer interaction. For more information, see https://yanmluk.github.io
\end{IEEEbiographynophoto}

\vspace{-35pt}

\begin{IEEEbiographynophoto}
{Akshith Reddy Kota} holds a MS in Computer Science, Stony Brook University. His research interests include machine learning, deep learning, explainable AI, visual analytics, and big data analytics. He earned a Bachelor of Technology in CSE from Vellore Institute of Technology, India.
\end{IEEEbiographynophoto}

\vspace{-35pt}

\begin{IEEEbiographynophoto}
{Eric Papenhausen} holds a PhD in Computer Science, Stony Brook University. His research interests includes machine learning, deep learning, explainable AI, visual analytics, big data analytics, and medical image synthesis.  He is currently CTO at Akai Kaeru LLC.
\end{IEEEbiographynophoto}

\vspace{-35pt}

\begin{IEEEbiographynophoto}
{Klaus Mueller} is currently a professor of Computer Science at Stony Brook University and a
senior scientist at Brookhaven National Lab. His
research interests include explainable AI, visual
analytics, data science, and medical imaging. He
won the US NSF Early CAREER Award and has co-authored $>$ 300 papers, cited more than 14,000 times.
He is a Fellow of the IEEE.
\end{IEEEbiographynophoto}

\end{document}